\documentclass{article}

\usepackage[nonatbib]{tackling_climate_workshop_style}

\usepackage[utf8]{inputenc} 
\usepackage[T1]{fontenc}    
\usepackage{hyperref}       
\usepackage{url}            
\usepackage{amsmath}
\usepackage{booktabs}       
\usepackage{amsfonts}       
\usepackage{nicefrac}       
\usepackage{microtype}      
\usepackage{graphicx}
\usepackage{multirow}
\usepackage{booktabs}
\usepackage{float}

\usepackage[backend=biber,style=numeric-comp]{biblatex}
\title{Spatially Regularized Graph Attention Autoencoder Framework for Detecting Rainfall Extremes}

\author{%
  Mihir Agarwal\thanks{Equal Contribution} , Progyan Das*, Udit Bhatia \\
  Indian Institute Of Technology Gandhinagar\\
  {\texttt{agarwalmihir}, 
  \texttt{progyan.das},
  \texttt{bhatia.u}}@iitgn.ac.in
}

\begin{document}

\maketitle

\begin{abstract}
We introduce a novel Graph Attention Autoencoder (GAE) with spatial regularization to address the challenge of scalable anomaly detection in spatiotemporal rainfall data across India from 1990 to 2015. Our model leverages a Graph Attention Network (GAT) to capture spatial dependencies and temporal dynamics in the data, further enhanced by a spatial regularization term ensuring geographic coherence. We construct two graph datasets employing rainfall, pressure, and temperature attributes from the Indian Meteorological Department and ERA5 Reanalysis on Single Levels, respectively. Our network operates on graph representations of the data, where nodes represent geographic locations, and edges, inferred through event synchronization, denote significant co-occurrences of rainfall events. Through extensive experiments, we demonstrate that our GAE effectively identifies anomalous rainfall patterns across the Indian landscape. Our work paves the way for sophisticated spatiotemporal anomaly detection methodologies in climate science, contributing to better climate change preparedness and response strategies.

\end{abstract}

\section{Introduction}

In computer vision, autoencoders have emerged as a versatile tool with a wide range of applications \cite{sinha2018deep} \cite{voulodimos2018deep}. Designed originally for unsupervised learning tasks \cite{baldi2012autoencoders}, these neural networks have demonstrated their effectiveness in capturing salient features from complex image data \cite{ghojogh2021generative}. By employing an encoder-decoder architecture, autoencoders aim to reconstruct their input as faithfully as possible after compressing it into a lower-dimensional latent space \cite{mathew2021deep}. This powerful capability makes them well-suited for various tasks such as image denoising \cite{singh2021overview} \cite{sawada2019review}, inpainting \cite{qin2021image} \cite{jam2021comprehensive}, feature extraction \cite{han2018autoencoder} \cite{liang2017text}, and anomaly detection \cite{baur2021autoencoders} \cite{li2022deep}. Their ability to learn data-driven transformations autonomously has rendered them an invaluable resource in handling the intricate challenges that often arise in visual computing.

In the realm of classical autoencoders, one of the prominent limitations lies in their inability to explicitly model long-range dependencies within the data \cite{ma2021comprehensive}. By their very design, conventional autoencoders generally concentrate on local feature extraction and reconstruction \cite{li2023comprehensive}, often neglecting the potentially critical long-range correlations that could exist within an image or between different data points. To surmount this limitation, Graph Autoencoders (GAEs) \cite{wu2020comprehensive} \cite{kim2022graph} leverage graph-based representations, allowing them to model both local and global data structures efficiently \cite{wu2020comprehensive}. The graph structure both encodes the immediate neighbors and facilitates connections to distant nodes, making it particularly suited for capturing phenomena that exhibit long-range teleconnections \cite{li2023graph}, such as extreme rainfall events \cite{steptoe2018correlations}.

In weather and climate studies, traditional autoencoders struggle with high-dimensional, spatially complex data, leading to computational inefficiency and potential overfitting. Graph Autoencoders (GAEs) offer a solution by encoding climate variables like precipitation and temperature into a scalable graph structure. This enables effective anomaly detection, accurate predictions, and denoising, while preserving crucial teleconnections often missed by conventional methods.

In this study, we introduce the \textbf{Spatially Regularized Graph Attention Autoencoder (SRGAttAE)}, a model designed to capture long-range spatial dependencies in climate variables for anomaly detection on rainfall events over India.

\section{Dataset}
We construct two datasets focused on the Indian subcontinent, spanning 26 years from 1990 to 2015. The dataset comprises daily graphs with 4827 nodes, each representing a unique geographical location. Approximately 75,000 to 85,000 edges are constructed through event synchronization. The dataset combines rainfall data from the Indian Meteorological Department (IMD) and rainfal from ERA5 single-level data respectively and other climate variables from ERA5 as the features. 

\textbf{Data:} 
The first dataset sources its rainfall measurements from the IMD \cite{mitra2013gridded}, while the other climate variable are obtained from ERA5 \cite{singh2021assessment} pressure-level Reanalysis data, which also provides additional variables like pressure, temperature, specific humidity etc.

\textbf{Graph Construction and Event Synchronization:} In our study, we formulate the edges of each yearly graph using Event Synchronization (ES). This method quantifies the temporal relationships between rainfall events at different geographical locations. The adjacency matrix, which captures the network's connectivity, is determined by both the strength and latency of these synchronized events and predefined thresholds. For a complete mathematical explanation, please refer to Appendix A.

\section{Methodology}
The core objective revolves around the detection of rainfall anomalies within the spatially-correlated data across the Indian subcontinent, represented as a graph structure. We propose the \textbf{SRGAttAE} to capture spatial dependencies effectively and ensure spatial coherence, which is quintessential for accurate anomaly detection in rainfall patterns \cite{zhou2017anomaly} \cite{cheng2021improved}.

\textbf{Model Architecture: }The SRGAttAE architecture is bifurcated into an encoding and a decoding phase, each equipped with two Graph Attention Network (GAT) layers \cite{veličković2018graph}. The encoding layers transition the input graph, passed as temporal snapshots to the model, consisting of 4827 nodes and 72,000-85,000 edges, into a 4-dimensional latent space. The decoding layers, on the other hand, reconstruct the original graph from this latent representation to minimize the reconstruction error, which is pivotal for anomaly detection \cite{zhao2017spatio}.

\begin{figure}[ht]
    \centering
    \hspace{-8mm}\includegraphics[scale = 0.14]{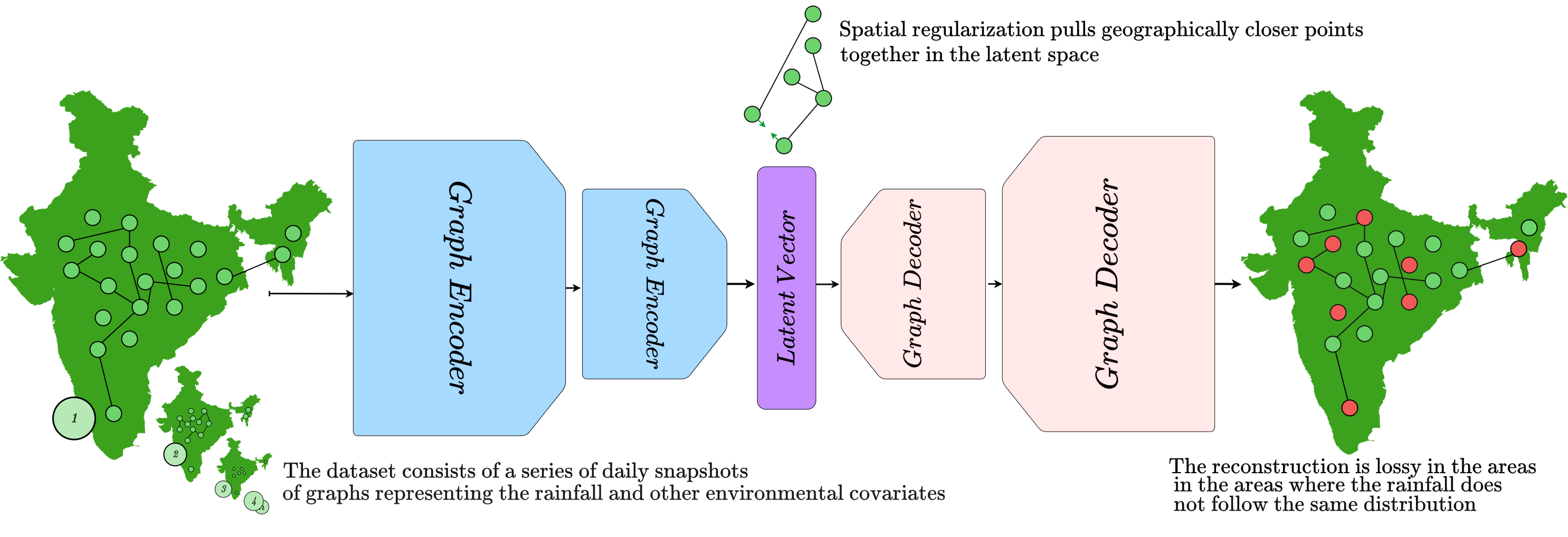}
    \caption{\textbf{SRGAttAE Model Architecture.} The data procurement and preprocessing are performed using publicly available IMD and ERA5 Reanalysis data.}
    \label{fig:SRGAttAE_arch}
\end{figure}

\textbf{Graph Attention Networks: }Graph Attention Networks (GAT) form the crux of our model, enabling the capture of spatial dependencies in the data. Unlike traditional convolutional layers \cite{kipf2017semisupervised}, GAT layers are designed to handle graph-structured data by leveraging self-attention mechanisms to weigh the importance of neighboring nodes dynamically. Formally, the attention coefficients between node \(i\) and node \(j\) are computed as follows \cite{veličković2018graph}:

\begin{equation}
    \alpha_{ij} = \frac{\exp\left(\text{LeakyReLU}\left(\mathbf{a}^T[\mathbf{W}\mathbf{x}_i \| \mathbf{W}\mathbf{x}_j]\right)\right)}{\sum_{k \in \mathcal{N}(i)} \exp\left(\text{LeakyReLU}\left(\mathbf{a}^T[\mathbf{W}\mathbf{x}_i \| \mathbf{W}\mathbf{x}_k]\right)\right)},
\end{equation}

where \(\mathbf{x}_i\) and \(\mathbf{x}_j\) are the feature vectors of nodes \(i\) and \(j\), \(\mathbf{W}\) is a shared linear transformation matrix, \(\mathbf{a}\) is the attention mechanism's learnable weight vector, and \(\mathcal{N}(i)\) denotes the neighbors of node \(i\). The architecture computes new node representations by aggregating the transformed features of neighboring nodes with attention-coefficients-modulated weights, enhancing the model's capacity to learn from the spatial correlations in the data \cite{zhang2019spatial}.

\textbf{Spatial Regularization: } The objective function of our \textit{SRGAttAE} model, \(\mathcal{L} = \mathcal{L}_{\text{Rec}} + \lambda \mathcal{L}_{\text{SCR}}\), balances graph reconstruction and spatial consistency. The Spatial Consistency Regularization (SCR) term, \(\mathcal{L}_{\text{SCR}} = \frac{1}{N}\sum_{i=1}^N \sum_{j=1}^N w_{ij} \left\| \mathbf{z}_i - \mathbf{z}_j \right\|^2\), ensures a spatially coherent embedding by minimizing the distance between latent representations of geographically neighboring nodes, where \(\lambda\) is a balancing hyperparameter and $w_{ij}$ is a Gaussian kernel based weight (refer to Appendix).

\textbf{Anomaly Detection:} Nodes manifesting high reconstruction error \cite{bank2021autoencoders} in the decoded graph are earmarked as potential anomalies. The reconstruction error serves as a robust indicator of rainfall anomalies, thus facilitating an effective anomaly detection mechanism across the spatial domain.

\section{Experiments and Results}

For our study, we train the \textit{SRGAttAE} model on five years of spatio-temporal rainfall data with spatial regularization. We evaluate its performance on 24 subsequent years of real-world data. The model excels in reconstructing general rainfall patterns but struggles with extreme values. A threshold, based on the $95^{\text{th}}$ percentile of daily reconstruction error, effectively identifies rainfall anomalies. Our results are consistent with traditional surveys, validating the model's utility.

\begin{figure}[ht]
    \centering
    \hspace{-19.25mm}
    \includegraphics[scale = 0.43]{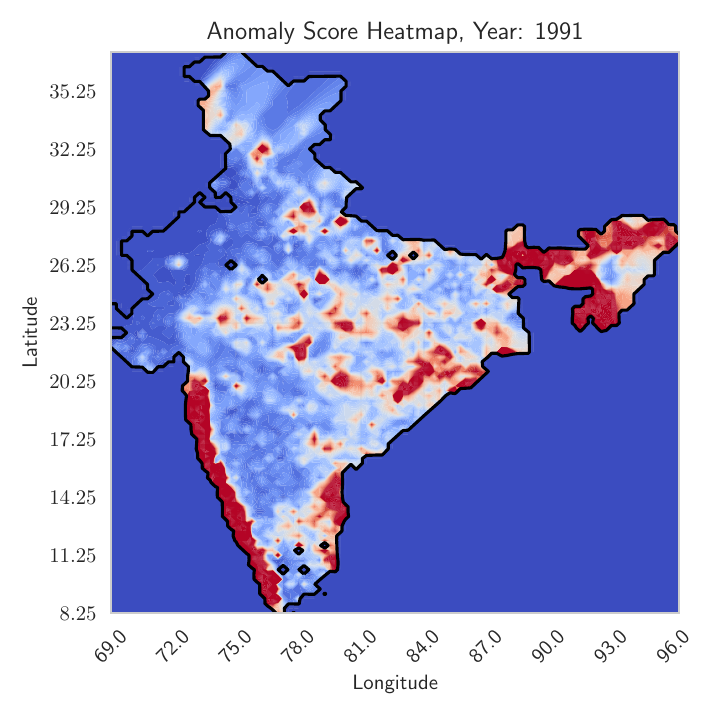}
    \includegraphics[scale = 0.43]{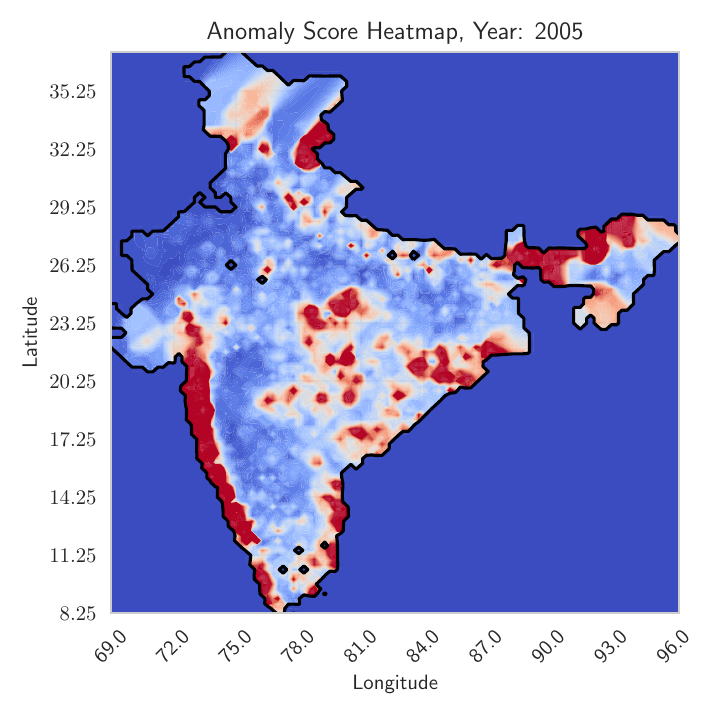}
    \includegraphics[scale = 0.43]{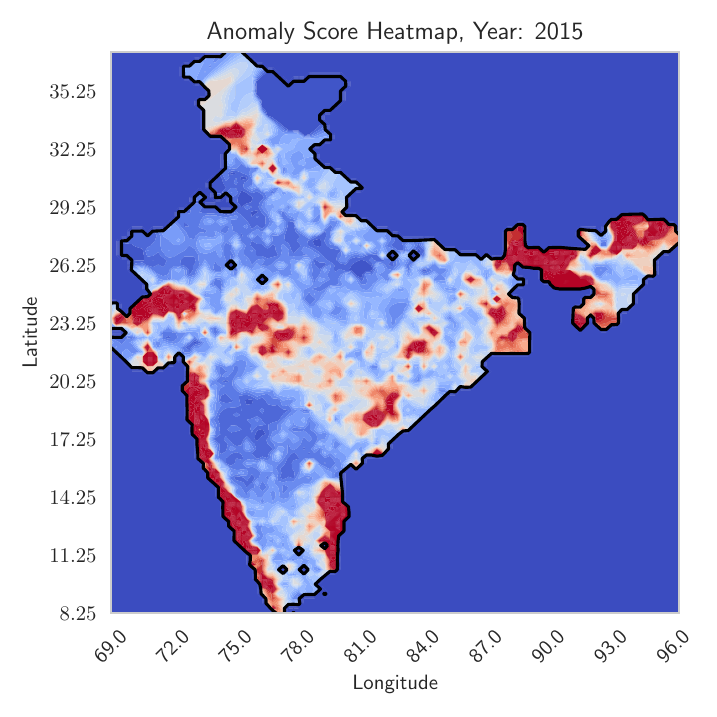}\\
        \hspace{-19.25mm}
    \includegraphics[scale = 0.40]{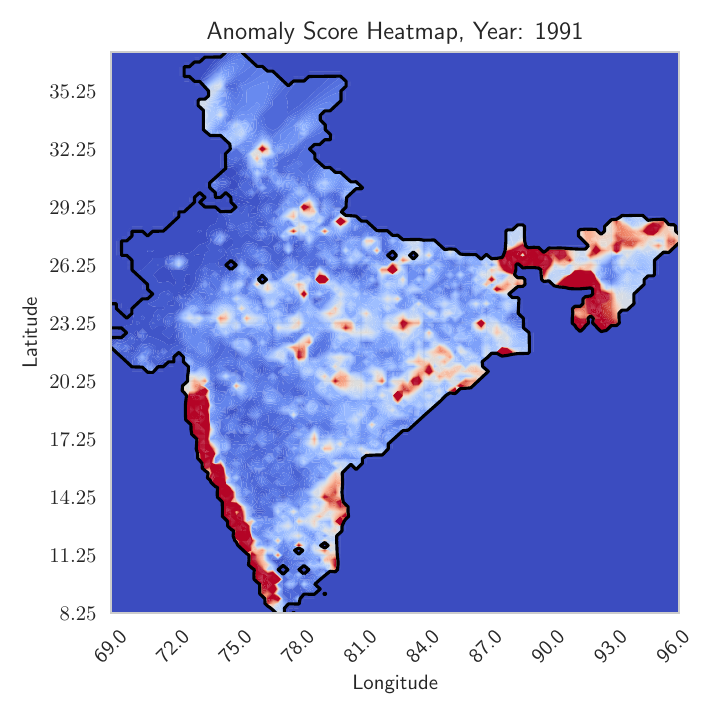}
    \includegraphics[scale = 0.40]{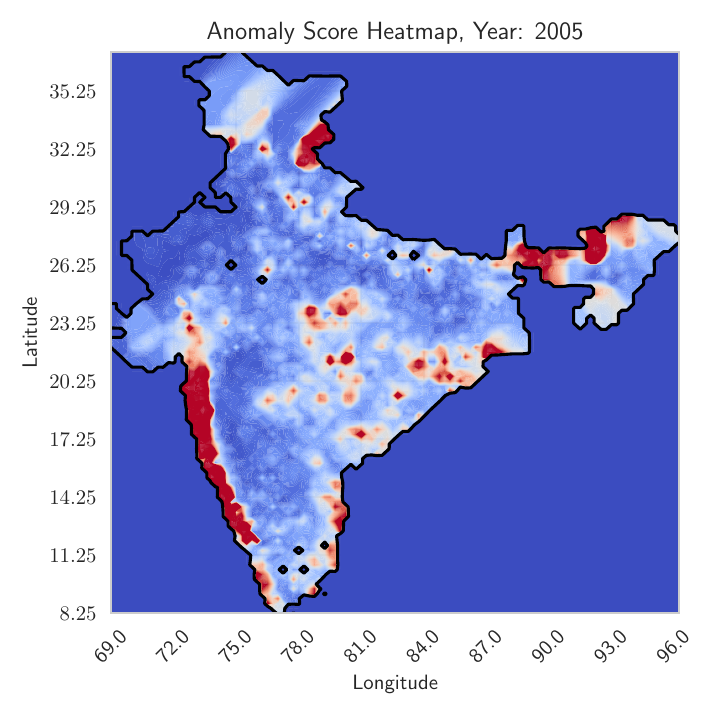}
    \includegraphics[scale = 0.40]{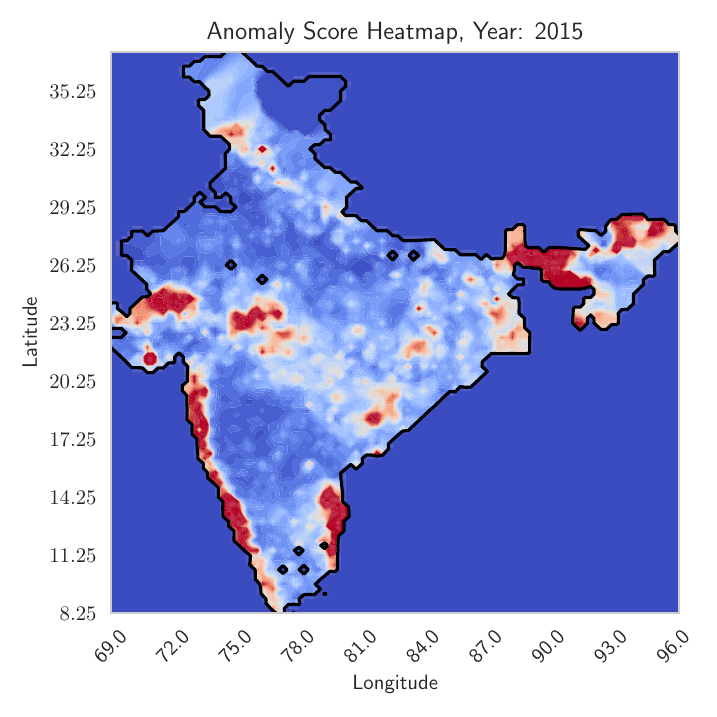}
    \caption{Heatmaps generated from the anomaly scores, for three years: 1991, 2005, and 2015. Uniformly red regions indicate $90^{th}$ percentile ($TOP$) and $95^{th}$ percentile ($BOTTOM$) anomaly scores. Heatmaps from 1991 through 2015 have been provided in the Appendix.}
    \label{fig:95_rain}
\end{figure}

Notably, our experiments are conducted on an Nvidia Quadro RTX 5000 GPU. Despite the complexity of the \textit{SRGAttAE} model, characterized by its attention-based architecture and spatial regularization features, the training phase concludes in approximately 6 minutes. This is particularly noteworthy given the substantial network size, which comprises $4,827$ nodes and fluctuates between $72,000$ and $85,000$ edges in the daily snapshots over the five-year training data. Inference operations spanning $25$ years are accomplished in just over one minute, underscoring the model's computational efficiency.

\textbf{Validation:} Our anomaly detection framework shows a high degree of corroboration with established surveys on extreme rainfall events. Importantly, rigorous experimentation substantiates that our identified anomalies are not merely instances of high rainfall across the Indian subcontinent. In an effort to refine our analysis, we employ a masking technique to exclude areas where the model overestimates rainfall. This focused approach narrows our evaluation to geographical regions that experience verifiably anomalous high rainfall events.

\begin{figure}[ht!]
    \centering
    \begin{minipage}{0.4\textwidth}
        \centering
        \begin{tabular}{lll}
            \toprule
            Model     & \multicolumn{2}{c}{Reconstruction} \\
            \cmidrule{2-3}
                       & MSE  & MAE \\
            \midrule
            GCN\footnote{GCN stands for Graph Convolutional Network, a non-attention based Graph Neural Network.\cite{kipf2016semi}.}  & 34.22  & 1.98  \\
            GCN with SCR     & 33.80   & 4.66    \\
            GAT & 32.10 & 1.78\\
            \textbf{GAT with SCR}     & \textbf{22.51} &   \textbf{1.46}    \\
            
            \midrule
            GCN & 22.52 & 1.73 \\
            GCN with SCR     & 23.10   & 1.69    \\
            GAT  & 19.02  & \textbf{1.43}  \\
            \textbf{GAT with SCR}     & \textbf{18.15} &   1.45    \\
            \bottomrule
        \end{tabular}
        \vspace{4mm}
        \caption{Experimentation Results on the IMD (top) and ERA5 (bottom) datasets; the MSE penalizes anomalies harsher than the MAE. As we can see, our GAT Autoencoder with SCR reconstructs the graph with the least or near-least error across both metrics.}
        \label{sample-table}
    \end{minipage}%
    \hfill
    \begin{minipage}{0.5\textwidth}
        \centering
        \includegraphics[width=\textwidth]{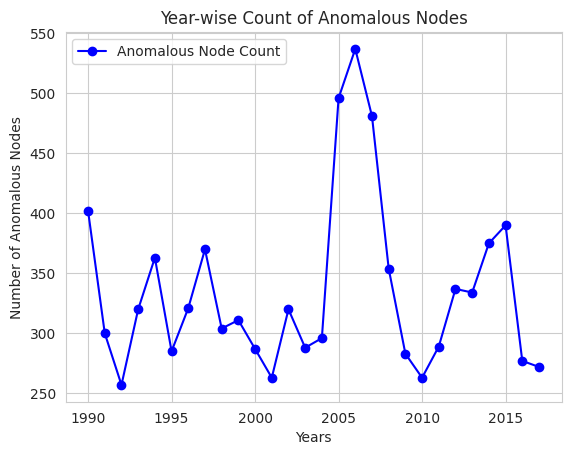}
        \caption{Variation of Node Anomalies over 1990-2015; statistical analysis using the Mann-Kendall trend test revealed no detectable trend. The temporal stability of rainfall events can be attributed to the enduring stability of the underlying graph structures, as shown in \cite{Tantary_2023}.}
        \label{fig:number}
    \end{minipage}
\end{figure}

\section{Conclusion}

In this study, we introduce the Spatially Regularized Graph Attention Autoencoder (SRGAttAE), a novel model designed for the unique climate complexities of the Indian subcontinent. The model excels in fast anomaly detection and shows broad applicability by generalizing well across multiple datasets. Our empirical analyses indicate stable anomaly patterns across the region with minor seasonal variations, offering valuable insights for stakeholders in climate science and setting the stage for future research in this critical field.

\printbibliography

@article{singh2021assessment,
  title={Assessment of newly-developed high resolution reanalyses (IMDAA, NGFS and ERA5) against rainfall observations for Indian region},
  author={Singh, Tarkeshwar and Saha, Upal and Prasad, VS and Gupta, M Das},
  journal={Atmospheric Research},
  volume={259},
  pages={105679},
  year={2021},
  publisher={Elsevier}
}

@article{mitra2013gridded,
  title={Gridded daily Indian monsoon rainfall for 14 seasons: Merged TRMM and IMD gauge analyzed values},
  author={Mitra, Ashis K and Momin, IM and Rajagopal, EN and Basu, S and Rajeevan, MN and Krishnamurti, TN},
  journal={Journal of Earth System Science},
  volume={122},
  pages={1173--1182},
  year={2013},
  publisher={Springer}
}

@article{qin2021image,
  title={Image inpainting based on deep learning: A review},
  author={Qin, Zhen and Zeng, Qingliang and Zong, Yixin and Xu, Fan},
  journal={Displays},
  volume={69},
  pages={102028},
  year={2021},
  publisher={Elsevier}
}

@article{li2022deep,
  title={Deep learning for anomaly detection in multivariate time series: Approaches, applications, and challenges},
  author={Li, Gen and Jung, Jason J},
  journal={Information Fusion},
  year={2022},
  publisher={Elsevier}
}

@article{li2023graph,
  title={Graph Neural Network for spatiotemporal data: methods and applications},
  author={Li, Yun and Yu, Dazhou and Liu, Zhenke and Zhang, Minxing and Gong, Xiaoyun and Zhao, Liang},
  journal={arXiv preprint arXiv:2306.00012},
  year={2023}
}

@article{steptoe2018correlations,
  title={Correlations between extreme atmospheric hazards and global teleconnections: Implications for multihazard resilience},
  author={Steptoe, H and Jones, SEO and Fox, H},
  journal={Reviews of Geophysics},
  volume={56},
  number={1},
  pages={50--78},
  year={2018},
  publisher={Wiley Online Library}
}

@article{li2023comprehensive,
  title={A comprehensive survey on design and application of autoencoder in deep learning},
  author={Li, Pengzhi and Pei, Yan and Li, Jianqiang},
  journal={Applied Soft Computing},
  pages={110176},
  year={2023},
  publisher={Elsevier}
}

@article{wu2020comprehensive,
  title={A comprehensive survey on graph neural networks},
  author={Wu, Zonghan and Pan, Shirui and Chen, Fengwen and Long, Guodong and Zhang, Chengqi and Philip, S Yu},
  journal={IEEE transactions on neural networks and learning systems},
  volume={32},
  number={1},
  pages={4--24},
  year={2020},
  publisher={IEEE}
}

@article{kim2022graph,
  title={Graph anomaly detection with graph neural networks: Current status and challenges},
  author={Kim, Hwan and Lee, Byung Suk and Shin, Won-Yong and Lim, Sungsu},
  journal={IEEE Access},
  year={2022},
  publisher={IEEE}
}

@article{ma2021comprehensive,
  title={A comprehensive survey on graph anomaly detection with deep learning},
  author={Ma, Xiaoxiao and Wu, Jia and Xue, Shan and Yang, Jian and Zhou, Chuan and Sheng, Quan Z and Xiong, Hui and Akoglu, Leman},
  journal={IEEE Transactions on Knowledge and Data Engineering},
  year={2021},
  publisher={IEEE}
}

@article{baur2021autoencoders,
  title={Autoencoders for unsupervised anomaly segmentation in brain MR images: a comparative study},
  author={Baur, Christoph and Denner, Stefan and Wiestler, Benedikt and Navab, Nassir and Albarqouni, Shadi},
  journal={Medical Image Analysis},
  volume={69},
  pages={101952},
  year={2021},
  publisher={Elsevier}
}

@article{liang2017text,
  title={Text feature extraction based on deep learning: a review},
  author={Liang, Hong and Sun, Xiao and Sun, Yunlei and Gao, Yuan},
  journal={EURASIP journal on wireless communications and networking},
  volume={2017},
  number={1},
  pages={1--12},
  year={2017},
  publisher={SpringerOpen}
}

@article{jam2021comprehensive,
  title={A comprehensive review of past and present image inpainting methods},
  author={Jam, Jireh and Kendrick, Connah and Walker, Kevin and Drouard, Vincent and Hsu, Jison Gee-Sern and Yap, Moi Hoon},
  journal={Computer vision and image understanding},
  volume={203},
  pages={103147},
  year={2021},
  publisher={Elsevier}
}

@article{sawada2019review,
  title={A review of blind source separation methods: two converging routes to ILRMA originating from ICA and NMF},
  author={Sawada, Hiroshi and Ono, Nobutaka and Kameoka, Hirokazu and Kitamura, Daichi and Saruwatari, Hiroshi},
  journal={APSIPA Transactions on Signal and Information Processing},
  volume={8},
  pages={e12},
  year={2019},
  publisher={Cambridge University Press}
}

@article{singh2021overview,
  title={An overview of variational autoencoders for source separation, finance, and bio-signal applications},
  author={Singh, Aman and Ogunfunmi, Tokunbo},
  journal={Entropy},
  volume={24},
  number={1},
  pages={55},
  year={2021},
  publisher={MDPI}
}

@inproceedings{baldi2012autoencoders,
  title={Autoencoders, unsupervised learning, and deep architectures},
  author={Baldi, Pierre},
  booktitle={Proceedings of ICML workshop on unsupervised and transfer learning},
  pages={37--49},
  year={2012},
  organization={JMLR Workshop and Conference Proceedings}
}

@article{ghojogh2021generative,
  title={Generative adversarial networks and adversarial autoencoders: Tutorial and survey},
  author={Ghojogh, Benyamin and Ghodsi, Ali and Karray, Fakhri and Crowley, Mark},
  journal={arXiv preprint arXiv:2111.13282},
  year={2021}
}

@article{mathew2021deep,
  title={Deep learning techniques: an overview},
  author={Mathew, Amitha and Amudha, P and Sivakumari, S},
  journal={Advanced Machine Learning Technologies and Applications: Proceedings of AMLTA 2020},
  pages={599--608},
  year={2021},
  publisher={Springer}
}

@article{sinha2018deep,
  title={Deep learning for computer vision tasks: a review},
  author={Sinha, Rajat Kumar and Pandey, Ruchi and Pattnaik, Rohan},
  journal={arXiv preprint arXiv:1804.03928},
  year={2018}
}

@inproceedings{han2018autoencoder,
  title={Autoencoder inspired unsupervised feature selection},
  author={Han, Kai and Wang, Yunhe and Zhang, Chao and Li, Chao and Xu, Chao},
  booktitle={2018 IEEE international conference on acoustics, speech and signal processing (ICASSP)},
  pages={2941--2945},
  year={2018},
  organization={IEEE}
}

@article{voulodimos2018deep,
  title={Deep learning for computer vision: A brief review},
  author={Voulodimos, Athanasios and Doulamis, Nikolaos and Doulamis, Anastasios and Protopapadakis, Eftychios and others},
  journal={Computational intelligence and neuroscience},
  volume={2018},
  year={2018},
  publisher={Hindawi}
}

@misc{veličković2018graph,
      title={Graph Attention Networks}, 
      author={Petar Veličković and Guillem Cucurull and Arantxa Casanova and Adriana Romero and Pietro Liò and Yoshua Bengio},
      year={2018},
      eprint={1710.10903},
      archivePrefix={arXiv},
      primaryClass={stat.ML}
}

@inproceedings{zhou2017anomaly,
  title={Anomaly detection with robust deep autoencoders},
  author={Zhou, Chong and Paffenroth, Randy C},
  booktitle={Proceedings of the 23rd ACM SIGKDD international conference on knowledge discovery and data mining},
  pages={665--674},
  year={2017}
}

@misc{kipf2017semisupervised,
      title={Semi-Supervised Classification with Graph Convolutional Networks}, 
      author={Thomas N. Kipf and Max Welling},
      year={2017},
      eprint={1609.02907},
      archivePrefix={arXiv},
      primaryClass={cs.LG}
}

@inproceedings{zhao2017spatio,
  title={Spatio-temporal autoencoder for video anomaly detection},
  author={Zhao, Yiru and Deng, Bing and Shen, Chen and Liu, Yao and Lu, Hongtao and Hua, Xian-Sheng},
  booktitle={Proceedings of the 25th ACM international conference on Multimedia},
  pages={1933--1941},
  year={2017}
}

@article{kipf2016semi,
  title={Semi-supervised classification with graph convolutional networks},
  author={Kipf, Thomas N and Welling, Max},
  journal={arXiv preprint arXiv:1609.02907},
  year={2016}
}

@article{zhang2019spatial,
  title={Spatial-temporal graph attention networks: A deep learning approach for traffic forecasting},
  author={Zhang, Chenhan and James, JQ and Liu, Yi},
  journal={IEEE Access},
  volume={7},
  pages={166246--166256},
  year={2019},
  publisher={IEEE}
}

@article{cheng2021improved,
  title={Improved autoencoder for unsupervised anomaly detection},
  author={Cheng, Zhen and Wang, Siwei and Zhang, Pei and Wang, Siqi and Liu, Xinwang and Zhu, En},
  journal={International Journal of Intelligent Systems},
  volume={36},
  number={12},
  pages={7103--7125},
  year={2021},
  publisher={Wiley Online Library}
}

@article{article,
author = {Stein, Michael},
year = {2005},
month = {02},
pages = {310-321},
title = {Space-Time Covariance Functions},
volume = {100},
journal = {Journal of the American Statistical Association},
doi = {10.1198/016214504000000854}
}

@misc{bank2021autoencoders,
      title={Autoencoders}, 
      author={Dor Bank and Noam Koenigstein and Raja Giryes},
      year={2021},
      eprint={2003.05991},
      archivePrefix={arXiv},
      primaryClass={cs.LG}
}

@article{Tantary_2023,
	doi = {10.22541/essoar.168677214.43959307/v1},
	url = {https://doi.org/10.22541%2Fessoar.168677214.43959307%2Fv1},
	year = 2023,
	month = {jun},
	publisher = {Authorea,
Inc.},
	author = {Danish Tantary and Arun K. Tangirala and Raghu Murtugudde and Rohini Kumar and Subimal Ghosh and Udit Bhatia},
	title = {Geographical Trapping of Synchronous Extremes amidst Increasing Variability of Indian Summer Monsoon Rainfall}
}

\appendix

\section{Appendix}
\subsection{Event Synchronisation}

Event synchronization is a powerful technique for assessing the correlation or coupling between discrete events in different time series. It is extensively employed in various domains, including climate science, neuroscience, and finance, due to its non-parametric nature and capability to capture intricate synchronization patterns. In the context of climate science, our study specifically focuses on the 4 summer monsoon months of JJAS (June, July, August, and September) and considers only those rain events that are above a certain percentile threshold for rainfall, thereby focusing on wet days.

\subsubsection{Mathematical Foundations}

The fundamental formulae involved in Event Synchronization are as follows:

\begin{equation}
s_{ij}^{lm} = \frac{1}{2}\min\{ \Delta t_{i}^{l}, \Delta t_{j}^{m} \},
\end{equation}

where \( \Delta t_{i}^{l} = |t_{i}^{l+1} - t_{i}^{l}| \) and \( \Delta t_{j}^{m} = |t_{j}^{m+1} - t_{j}^{m}| \).

The occurrences of events at \( i \) after \( j \) and vice versa are governed by \( c(i|j) \), defined as:

\begin{equation}
c(i|j) = \sum_{l,m} J_{ij} \quad \text{where} \quad J_{ij} =
\begin{cases} 
1 & \text{if } 0 \leq \Delta t_{lm}^{ij} \leq s_{ij}^{lm}, \\
1/2 & \text{if } \Delta t_{lm}^{ij} = 0, \\
0 & \text{otherwise},
\end{cases}
\end{equation}

The synchronization strength and delay, \( Q_{ij} \), are then given by:

\begin{equation}
Q_{ij} = \frac{c(i|j) + c(j|i)}{\sqrt{s_{ij}}}.
\end{equation}

Finally, the adjacency matrix \( A \) for the network is derived as:

\begin{equation}
A_{ij} = 
\begin{cases} 
1 & \text{if } Q_{ij} > \theta_{Q_{ij}}, \\
0 & \text{otherwise}.
\end{cases}
\end{equation}
\begin{figure}[H]
    \centering
    \hspace{-10mm}\includegraphics[scale = 0.35]{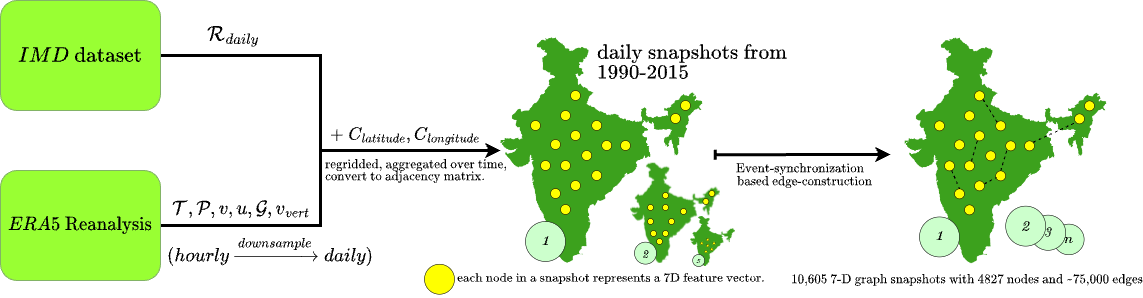}
    \caption{A schematic of the dataset construction process.}
    \label{fig:enter-label}
\end{figure}
\subsubsection{Data Considerations}

In our study, we employ daily, gridded rainfall data from 1990 to 2015, focusing specifically on the Indian region during the monsoon months of June, July, August and September (JJAS). We consider only those rain events that are above 1 mm rainfall, thereby allowing us to concentrate on wet days. This targeted approach enhances the robustness and applicability of our findings to monsoon-related climate studies.

\section{Experimentation of ERA 5 rainfall Dataset}

In our experiments, we also utilized the ERA5 single-level reanalysis dataset to assess the \textbf{SRGAttAE} model's adaptability and robustness. Employing the same experimental setup as with the Indian Meteorological Department (IMD) dataset, our model maintained its superior performance, achieving a mean square error (MSE) of 18.15 and a mean absolute error (MAE) of 1.45. These results, documented in Table 1 of the appendix, underline the model's capability to generalize across diverse climate datasets.

\begin{table}[H]
    \centering
    \begin{tabular}{lll}
        \toprule
        Model     & \multicolumn{2}{c}{Reconstruction} \\
        \cmidrule{2-3}
                  & MSE  & MAE \\
        \midrule
        GCN & 22.52 & 1.73 \\
        GCN with SCR     & 23.10   & 1.69    \\
        GAT  & 19.02  & \textbf{1.43}  \\
        \textbf{GAT with SCR}     & \textbf{18.15} &   1.45    \\
        \bottomrule
    \end{tabular}
    \caption{Benchmarks over the ERA5 dataset}
    \label{your_label_here_i}
\end{table}

\section{Graph Attention Autoencoder (GAE)}
Our proposed Graph Attention Autoencoder (GAE) consists of an encoder and a decoder, both built upon the Graph Attention Network (GAT) framework. The encoder compresses the input graph into a lower-dimensional latent space, while the decoder reconstructs the original graph from the latent representation. The architecture is formulated as follows:

The encoding phase aims to learn a lower-dimensional representation of the input graph while preserving its spatial structure. It comprises two Graph Attention (GAT) layers. The first GAT layer, $\text{GATConv}_1$, takes the rainfall measurement at each node, concatenated with its geographic location, as input, and outputs a 256-dimensional feature vector. The second GAT layer, $\text{GATConv}_2$, further processes these features into a 64-dimensional latent representation encapsulating the spatial relationships among nodes. Mathematically, the encoding phase is represented as:
\begin{align}
    \mathbf{h}_1 &= \text{ReLU}(\text{GATConv}_1(\mathbf{X}, \mathbf{E})), \\
    \mathbf{z} &= \text{ReLU}(\text{GATConv}_2(\mathbf{h}_1, \mathbf{E})),
\end{align}
where $\mathbf{X}$ denotes the node features, $\mathbf{E}$ denotes the edges, and $\mathbf{z}$ is the latent representation.

The decoding phase aims to reconstruct the original graph from the latent representation $\mathbf{z}$. It also comprises two GAT layers, $\text{GATConv}_3$ and $\text{GATConv}_4$, which attempt to invert the encoding process and recover the original node features. The decoding phase is represented as:
\begin{align}
    \mathbf{h}_2 &= \text{ReLU}(\text{GATConv}_3(\mathbf{z}, \mathbf{E})), \\
    \hat{\mathbf{X}} &= \text{GATConv}_4(\mathbf{h}_2, \mathbf{E}),
\end{align}
where $\hat{\mathbf{X}}$ is the reconstructed node features.

\noindent where \(\mathbf{X}\) is the input graph, \(\mathbf{A}\) is the adjacency matrix, \(\mathbf{Z}\) is the latent representation, and \(\hat{\mathbf{X}}\) is the reconstructed graph.

\subsection{Spatial Regularization}
To ensure geographic coherence in the latent space, we introduce a spatial regularization term, \(\mathcal{R}_{\text{spatial}}\), into the loss function. This term encourages nodes closer in geographic space to have similar latent representations. It is defined as:

\begin{align}
    \mathcal{R}_{\text{spatial}} = \sum_{i, j} w_{ij} \left\| \mathbf{z}_i - \mathbf{z}_j \right\|^2,
\end{align}

\noindent where \(w_{ij} = \exp\left(-\frac{d_{ij}^2}{2\sigma^2}\right)\) is a Gaussian kernel with \(d_{ij}\) being the geographic distance between nodes \(i\) and \(j\), and \(\sigma\) is a bandwidth parameter.

\subsection{Training Criterion}
The training objective is to minimize a combination of the reconstruction loss, \(\mathcal{L}_{\text{recon}}\), and the spatial regularization loss, \(\mathcal{L}_{\text{spatial}}\), defined as follows:

\begin{align}
    \mathcal{L}_{\text{recon}} &= \left\| \mathbf{X} - \hat{\mathbf{X}} \right\|^2_F, \\
    \mathcal{L}_{\text{spatial}} &= \lambda \cdot \mathcal{R}_{\text{spatial}}, \\
    \mathcal{L}_{\text{total}} &= \mathcal{L}_{\text{recon}} + \mathcal{L}_{\text{spatial}},
\end{align}

\noindent where \(\left\| \cdot \right\|^2_F\) denotes the Frobenius norm, and \(\lambda\) is a hyperparameter controlling the trade-off between reconstruction and spatial regularization. The model is trained by minimizing \(\mathcal{L}_{\text{total}}\) using stochastic gradient descent (SGD) or other suitable optimization algorithms.

\end{document}